\documentclass[runningheads,a4paper]{llncs}
\usepackage{amssymb}
\usepackage{graphicx}
\usepackage{times}

\begin{document}

\pagestyle{empty}

\mainmatter

\title{Clustering of Local Optima\\ in Combinatorial Fitness Landscapes}

\titlerunning{Clustering of Local Optima in Combinatorial Fitness Landscapes}

\author{Gabriela Ochoa\inst{1}, Sebastien V\'erel\inst{2},  Fabio Daolio\inst{3}, Marco Tomassini\inst{3}}

\authorrunning{G. Ochoa,  S. V\'erel, M. Tomassini and F. Daolio}   

\institute{School of Computer Science, University of Nottingham, Nottingham, UK. \and INRIA Lille - Nord Europe and University of Nice Sophia-Antipolis, France. \and Information Systems Department, University of Lausanne, Lausanne, Switzerland. }

\maketitle

\begin{abstract}
Using the recently proposed model of combinatorial landscapes: {\em local optima networks}, we study the distribution  of local optima in two  classes of  instances of the {\em quadratic assignment problem}. Our results indicate that the two problem instance classes give rise to very different configuration spaces. For the so-called real-like class,  the optima networks possess a clear modular structure, while the networks belonging to the class of random uniform instances are less well partitionable into clusters. We briefly discuss  the consequences of the findings for heuristically searching the corresponding problem spaces.
\end{abstract}

\section{Introduction}
We have recently introduced  a  model of combinatorial landscapes: \textit{Local Optima Networks} (LON) ~\cite{pre09,ieee-neutral}, which  allows the use of complex network analysis techniques~\cite{doye02} for  studying  fitness landscapes and problem difficulty in combinatorial optimization. The model, inspired by work in the physical sciences on energy surfaces\cite{newman03}, is based on the idea of compressing the information given by the whole problem configuration space into a smaller mathematical object which is the graph having as vertices the local optima  and as edges the possible transitions between them. This characterization of landscapes as networks has brought new insights into the global structure of the landscapes studied.  Moreover, some network features have been found to correlate and suggest explanations for search difficulty on the studied domains. Our initial work considered binary search spaces and the NK family of abstract landscapes~\cite{pre09,ieee-neutral}. Recently, we have turned our attention to more realistic combinatorial spaces (permutation spaces), specifically,  the \textit{Quadratic Assignment Problem} (QAP)~\cite{QAP-cec-10}. In this article,  we focus on a particular characteristic of the optima networks using the QAP, namely, the manner in which local optima are distributed in the configuration space. Several questions can be raised. Are they uniformly distributed, or do they cluster in some non-homogeneous way? If the latter, what is the relation between objective function values within and among different clusters and how easy is it to go from one cluster to another? Knowing even approximate answers to some of these questions would be very useful to further characterize the difficulty of a class of problems and also, potentially, to devise new search heuristics or variation to known heuristics that take advantage of this information. This short paper starts to address some of these questions. The sections below summarize our methodology and preliminary results.

\section{Methodology}
\subsection{The Quadratic Assignment Problem}
The QAP is a combinatorial problem in which a set of facilities with given flows has to be assigned to a set of locations with given distances in such a way that the sum of the product of flows and distances is minimized. A solution to the QAP is generally written as a permutation $\pi$ of the set $\{1,2,...,n\}$. The cost associated with a permutation $\pi$ is: $C(\pi)=\sum_{i=1}^{n}\sum_{j=1}^{n}{a_{ij}b_{\pi_{i}\pi_{j}}}$,  where $n$ denotes the number of facilities/locations and  $A=\{a_{ij}\}$ and $B=\{b_{ij}\}$ are referred to as the distance and flow matrices, respectively. The structure of these two matrices  characterizes the class of instances of the QAP problem.
For the statistical analysis conducted here,  the two instance generators proposed in~\cite{Knowles2003emo} for the multi-objective QAP were adapted for the single-objective QAP. The first generator produces uniformly random instances where all flows and distances are integers sampled from uniform distributions.  The second generator produces flow entries that are non-uniform random values. The instances produced have the so called ``real-like'' structure since they resemble the structure of QAP problems found in practical applications. For the purpose of community detection, $200$  instances were produced and analyzed with size $9$ for the random uniform class, and $200$ of size $11$ for the real-like instances class. Problem size $11$ is the largest one for which an exhaustive sample of the configuration space was computationally feasible in our implementation.

\subsection{Local Optima Networks}

In order to define the local optima network of the QAP instances, we need to provide the definitions for the nodes and  edges of the network. The vertexes of the graph can be straightforwardly defined as the local minima of the landscape, w.r.t. the neighborhood defined by a $2$-opt swap in the permutation space. In this work,  we select small QAP instances such that it is feasible to obtain all the nodes exhaustively by running a best-improvement hill-climbing algorithm from every configuration of the search space. 
The edges account for the transition probability between basins of attraction of the local optima. More formally, the edges reflect the total probability of going from basin $b_i$ to basin $b_j$, which is the average over all $s \in b_i$ of the transition probabilities  to solutions $s^{'} \in b_j$.
The reader is referred to~\cite{QAP-cec-10} for a more detailed exposition.

\noindent We define a \textit{Local Optima Network} (LON) as being the graph  $G=(S^*,E)$ where the set of vertices $S^*$ contains all the local optima, and there is an edge $e_{ij} \in E$ with weight $\vec{w}_{ij} = p(b_i \rightarrow b_j)$ between two nodes $i$ and $j$ iff $p(b_i \rightarrow b_j) > 0$. Notice that since each maximum has its associated basin, $G$ also describes the interconnection of basins.

The study of LONs for the QAP instances~\cite{QAP-cec-10}, showed that the networks are  dense. Indeed, they are complete or almost complete graphs,  which  is inconvenient for cluster detection algorithms. Moreover, from the perspective of heuristic search algorithms, only the most likely transitions play a role. Therefore, we opted for filtering out the networks edges keeping the more likely transitions. In filtering, we first replace the directed graph by an undirected one ($w_{ij}= \frac{\vec{w}_{ij} + \vec{w}_{ji}}{2}$), and then suppress all edges that have $w_{ij}$ smaller than the value making the $\alpha$-quantile ($\alpha=0.05$ in experiments)
in the weights distribution. Such a less dense network provides a coarser but clearer view of the fitness landscape backbone, and can be used for minima cluster analysis.

\section{Results and Discussion}

\begin{figure}[h!]
\begin{center}
 \includegraphics[width=0.55\textwidth]{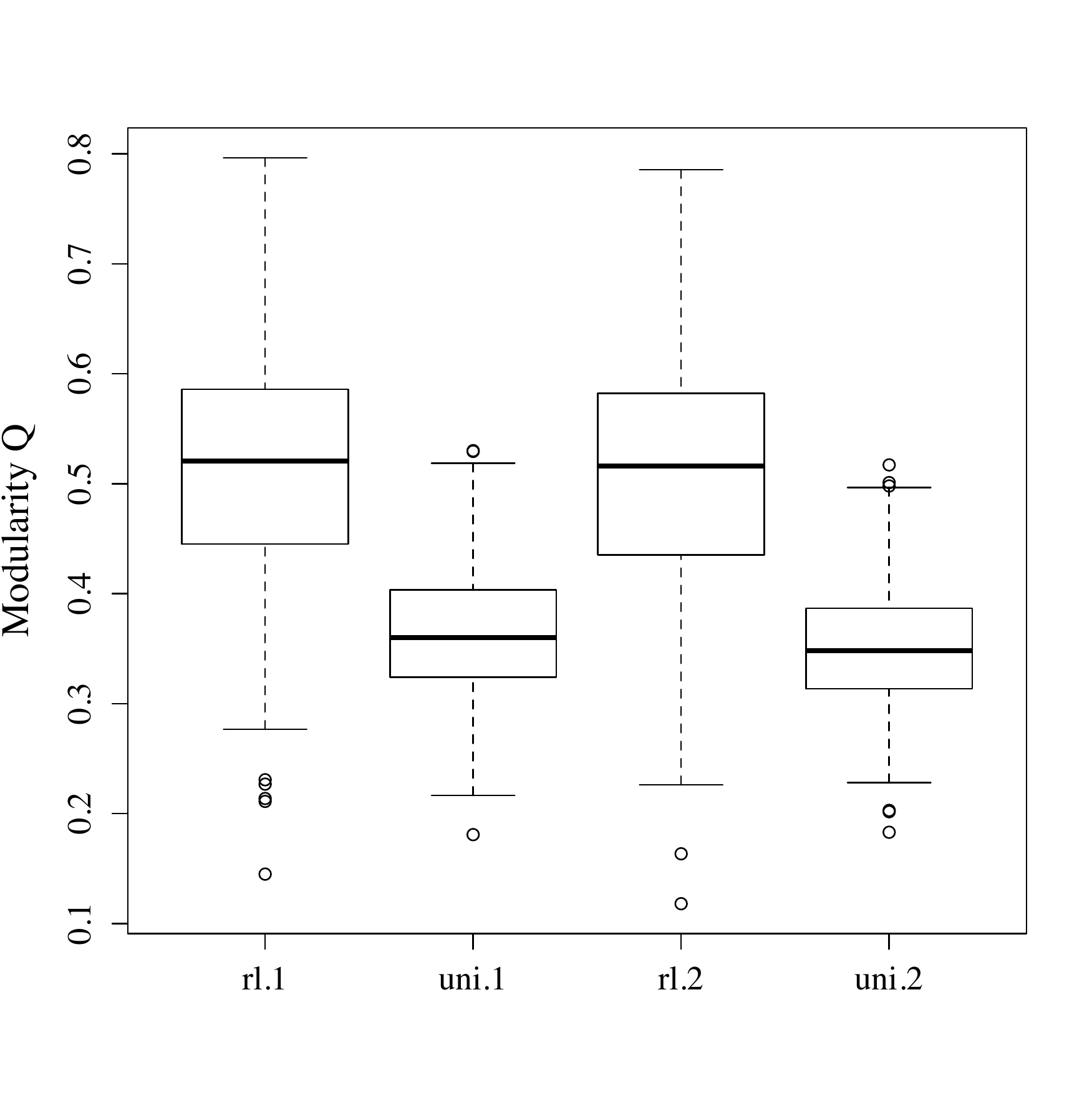}
  \vspace{-0.4cm}
 \caption{Boxplots of the modularity score $Q$ on the y-axis with respect to class problem (rl stands for real-like and
  uni stands for random uniform) and community detection algorithm (1 stands for fast greedy modularity
  optimization and 2 stands for spin glass search algorithm).
  }
\label{fig:box}
\end{center}
\end{figure}

Clusters or communities in networks can be loosely defined as being groups of nodes that are strongly  connected between them and poorly connected with the rest of the graph.  Community detection is a difficult task, but today several good approximate algorithms are available~\cite{santo1}. Here we use two of them: (i) a method based on greedy modularity optimization, and (ii) a spin glass ground state-based algorithm, in order to double check the community partition results. Figure~\ref{fig:box} shows the modularity score ($Q$) distribution calculated for each algorithm/instance-class.  In general, the higher the value of $Q$ of a partition, the crisper the community structure~\cite{santo1}.  The plot indicates that the two instance classes are well separated in terms of $Q$, and that the community detection algorithm does not seem to have any influence on such a result.

\begin{figure}[h!]
\begin{center}
 \includegraphics[width=0.47\textwidth]{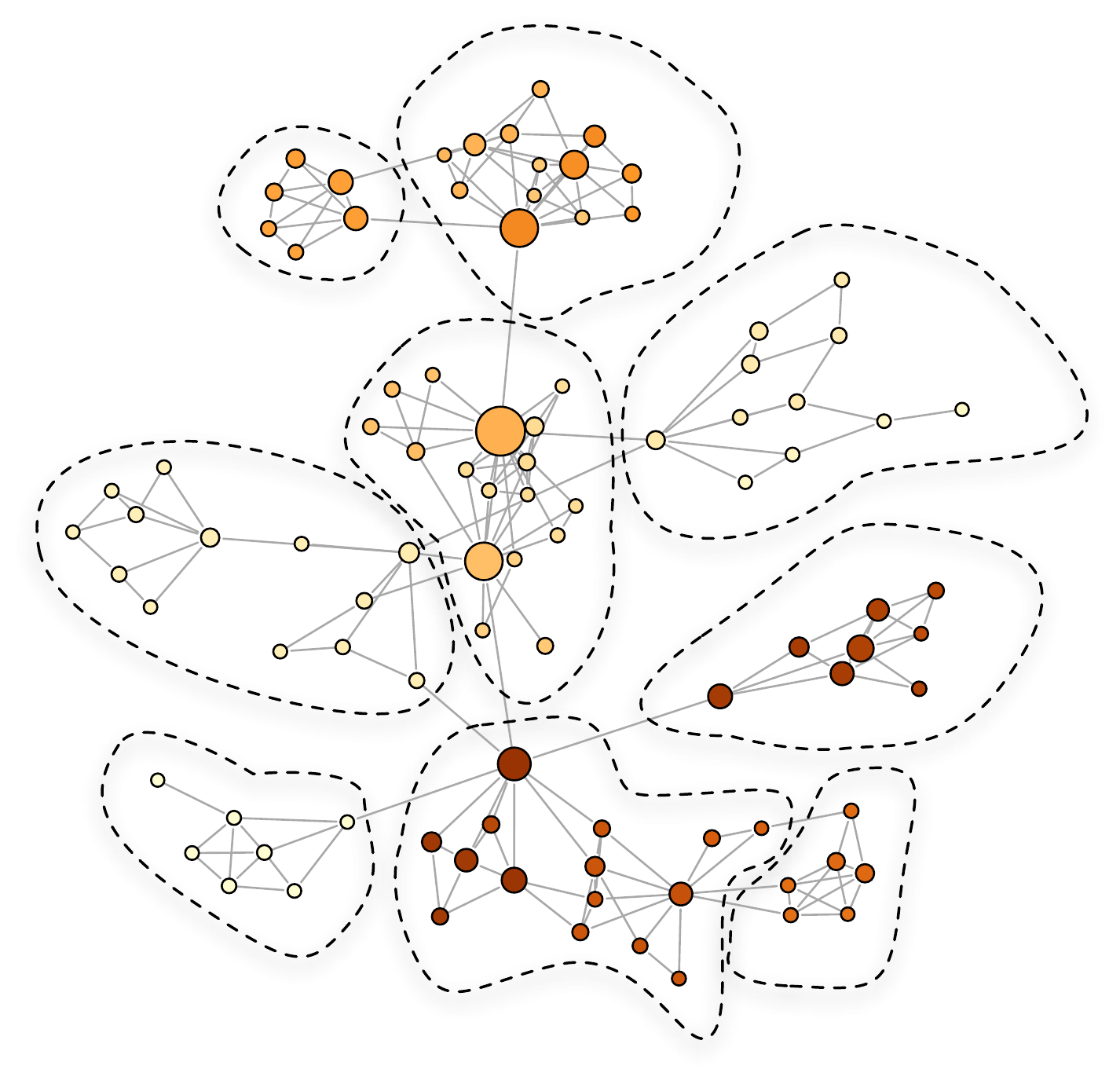}
 \includegraphics[width=0.47\textwidth]{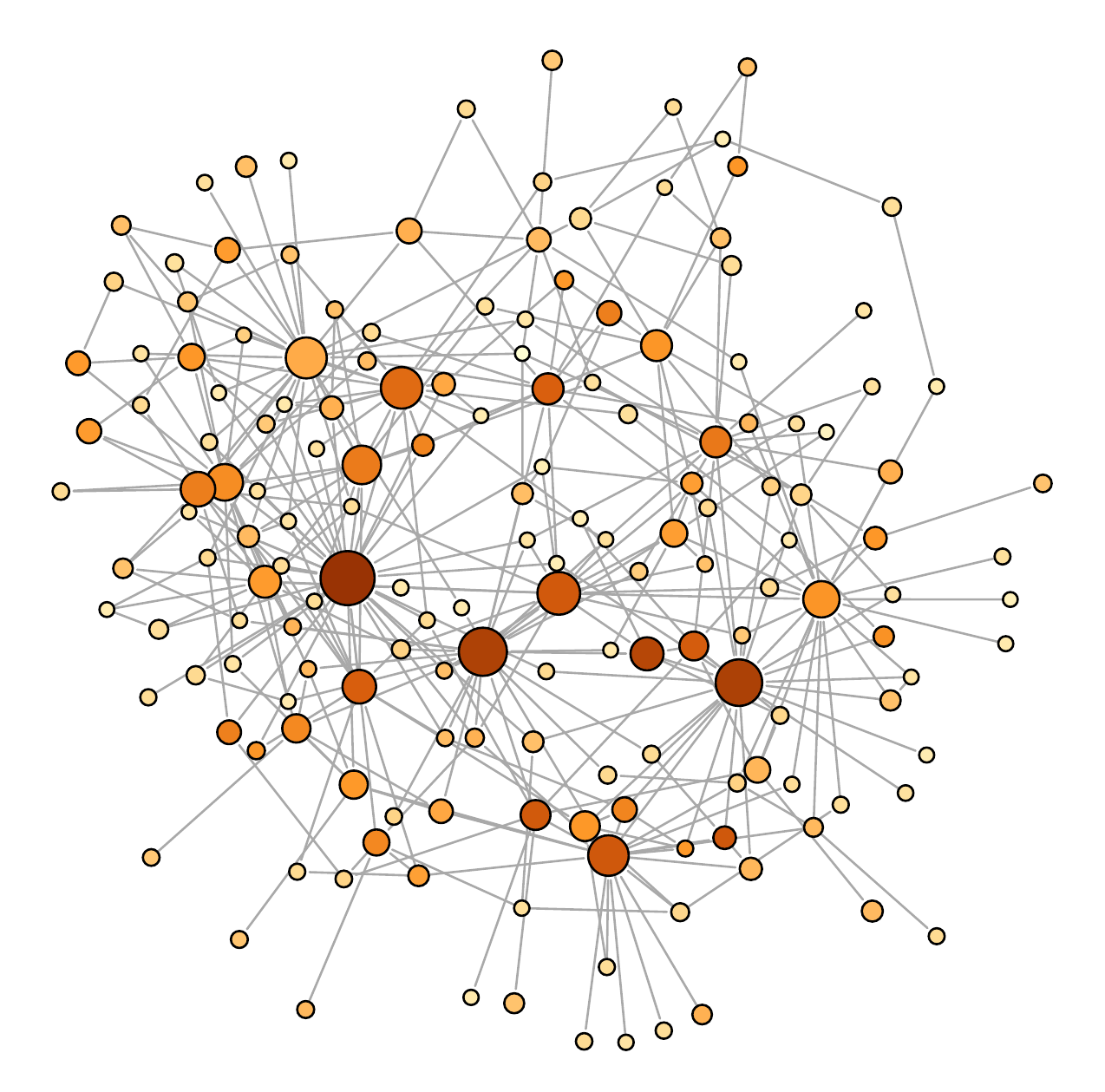}
  \vspace{-0.4cm}
 \caption{Community structure of the filtered LONs for two selected instances: real-like (Left); uniform (Right). Node sizes are proportional to the corresponding basin size. Darker colors mean better fitness. The layout has been produced with the R interface to the igraph library.}
\label{fig:comm}
\end{center}
\end{figure}

The modularity measurements (Fig. \ref{fig:box}) indicate  that real-like instances have significantly more minima cluster structure than the class of random uniform instances of the QAP problem. This can be appreciated visually by looking at Fig.~\ref{fig:comm} where the community structures of the LON of two particular instances are depicted. Although these are the two particular cases with the highest $Q$ values of their respective classes, the trends observed are general. For the real-like instance (Fig.~\ref{fig:comm}, left) one can see that groups of minima are rather recognizable and form well separated clusters (encircled with dotted lines), which is also reflected in the high corresponding modularity value $Q=0.79$. Contrastingly, the right plot represents a case drawn from the class of random uniform instances. The network has communities, with a $Q=0.53$, although they are hard to represent graphically, and thus are not shown in the picture.

The clustering results discussed above may have deep consequences on the heuristic algorithms used to search the corresponding landscapes. For example, on the  random uniform instances a simple local heuristic search, such as hill-climbing, should be sufficient to quickly find satisfactory solutions since they are homogeneously distributed. In contrast, in the real-like case they are much more clustered in regions of the search space. This leads to more modular optima networks and using multiple parallel searches, or large neighborhood moves would probably be good strategies. These ideas clearly deserve further investigation.  Future work will confirm the statistical significance of our results, consider larger instances using appropriate sampling, and  explore additional combinatorial problems.

  \vspace{-0.5cm}
\small
\bibliographystyle{splncs03}

\end{document}